\documentclass{svproc}
\usepackage{url}
\usepackage{graphicx}
\usepackage{multicol}
\usepackage{multirow}
\usepackage{footmisc}
\usepackage{amsmath}
\usepackage{amssymb}
\usepackage{siunitx}
\usepackage{booktabs}
\usepackage[export]{adjustbox}
\usepackage{tikz}
\usepackage{pgfplots}
\usepackage{svg}
\usepackage{algorithm}
\usepackage{algpseudocode}
\usepgfplotslibrary{fillbetween}
\pgfplotsset{compat=newest}

\usepackage[caption=false,font=footnotesize]{subfig}

\newcommand{\speciallabel}[2]{
  \edef\@currentlabel{#1}\label{#2}%
}

\usepackage{enumerate}

\DeclareMathOperator*{\argmin}{\arg\!\min}

\begin{document}
\mainmatter              

\title{On the Benefits of Robot Platooning \\
for Navigating Crowded Environments}
\titlerunning{On the Benefits of Robot Platooning}  
%
\author{Jahir Argote-Gerald\inst{1} \and Genki Miyauchi\inst{1} \and
Paul Trodden\inst{1} \and Roderich Gro\ss\inst{1,2}}
\authorrunning{J. Argote-Gerald et al.} 
%
%
\institute{School of Electrical and Electronic Engineering, The University of Sheffield, Sheffield, UK,\\
\email{\{jaargotegerald1,g.miyauchi,p.trodden\}@sheffield.ac.uk}
\and
Department of Computer Science, Technische Universität Darmstadt, Darmstadt, Germany,\\
\email{{roderich.gross@tu-darmstadt.de}}
}

\maketitle              

\newcommand{\cmt}[1]{{\color{red} #1}}
\newcommand{\cmtt}[1]{{\color{blue} #1}}

\begin{abstract}
This paper studies how groups of robots can effectively navigate through a crowd of agents. It quantifies the performance of platooning and less constrained, greedy strategies, and the extent to which these strategies disrupt the crowd agents. Three scenarios are considered: (i) passive crowds, (ii) counter-flow crowds, and (iii) perpendicular-flow crowds. 
Through simulations consisting of up to 200 robots, we show that 
for navigating passive and counter-flow crowds, the platooning strategy is less disruptive and more effective in dense crowds than the greedy strategy, whereas for navigating perpendicular-flow crowds, the greedy strategy outperforms the platooning strategy in either aspect.
Moreover, we propose an adaptive strategy that can switch between platooning and greedy behavioral states, and demonstrate that it combines the strengths of both strategies 
in all the scenarios considered.
\end{abstract}


\section{Introduction}
\label{Intro}


A platoon refers to a group of agents that move together in leader-follower pairs while maintaining visibility~\cite{morbidi2011visibility}. Research on platooning initially focused on vehicle-following applications in single- or multi-lane road networks~\cite{goli2019mpc}. The use of platoons in these applications can be motivated by potential reductions in costs and emissions~\cite{pi2023automotive}, and enhanced highway safety and efficiency~\cite{li2015overview}. Recent research has demonstrated platooning in less constrained 2-D environments, for example with ground-based~\cite{cruz2018chain,hirata2021leader} and water-based~\cite{hu2023spontaneous} robots.

Platooning requires a leader-follower formation in which each agent interacts with at most two other agents. 
This can be beneficial as it simplifies (i) the communication topology among the agents and (ii) the design and implementation for a decentralized controller~\cite{chen2023survey}.
However, it 
also restricts the agent's motion, which could compromise performance or even the group's ability to navigate complex environments.
Thus, it is of interest to explore whether conditions exist where platoon formations perform on par with, or even superior to, less constrained control strategies.

We are particularly interested in the ability of a group of robots to navigate environments populated with dynamic obstacles.
In many applications, robots may have to navigate through a human crowd, for example, when operating in shopping malls and plazas~\cite{truong2017toward,luo2018porca}, factories~\cite{mavrogiannis2019effects} or warehouses~\cite{fan2020distributed}. 
Prior works have considered incorporating social rules into the robot's controller~\cite{yang2019social}, using a robot to modify the crowd behavior to assist in evacuations~\cite{boukas2015robot}, and performing ``interaction actions'' to navigate unstructured crowds~\cite{dugas2020ian}.
However, these works consider only a single robot navigating through densely populated human crowds.
To the best of our knowledge, ours is the first study that quantifies the ability of a group of robots to collectively navigate through dense crowds.

When two crowds encounter each other while moving in opposite~\cite{feliciani2016empirical} or perpendicular~\cite{bacik2023lane} directions, lanes spontaneously form between them. This lane formation phenomenon occurs not only in human crowds but also at the molecular level~\cite{reichhardt2018velocity}. In general, it is an open problem whether such lane formations could benefit from agents that are explicitly programmed to follow each other.

In this paper, we consider the problem for a group of robots to navigate through a dynamic crowd. 
We examine the effectiveness of a platooning strategy, and a less constrained, greedy strategy,
as well as the extent of disruption they cause to 
the agents of the crowd. For either strategy, a distributed controller is presented.
We validate the controllers using three scenarios.
The results show that for navigating passive and counter-flow crowds, the platooning strategy is less disruptive, and where the crowds are dense, it is also more effective than the greedy strategy, whereas for navigating perpendicular-flow crowds, the greedy strategy outperforms the platooning strategy in either aspect.
Moreover, we propose an adaptive variant of the platooning controller in which platoon formations can dynamically split or merge, and show this to mitigate the shortcomings of either platooning or greedy strategies.

The paper is organized as follows. 
Section~\ref{PF} describes the problem formulation. Section~\ref{CS} presents the control strategies. Section~\ref{SS} describes the simulation setup. 
Section~\ref{MI} provides insights gained when using the social force model in our context.
Section~\ref{RD} presents the results. Finally, Section~\ref{Conc} concludes the paper.

\section{Problem Formulation}
\label{PF}

\begin{figure}[t]
    \centering
    \begin{tabular}{cc}
    \multirow{2}{*}{
    \subfloat[Passive crowd]
    {\includegraphics[width=0.49\linewidth,valign=m]{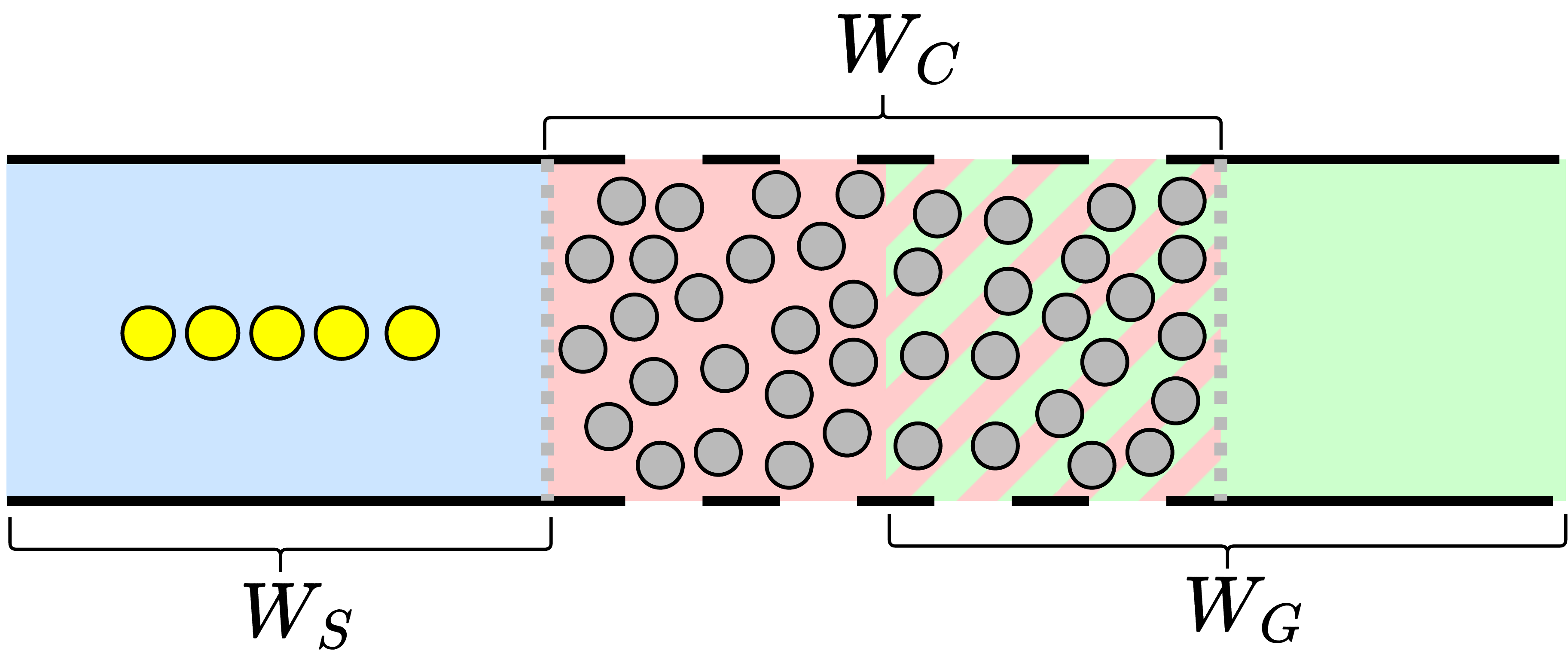}
    \label{fig:env_scenarios:no-flow}}} & {\subfloat[Counter-flow crowd]{\includegraphics[width=0.49\linewidth,valign=m]{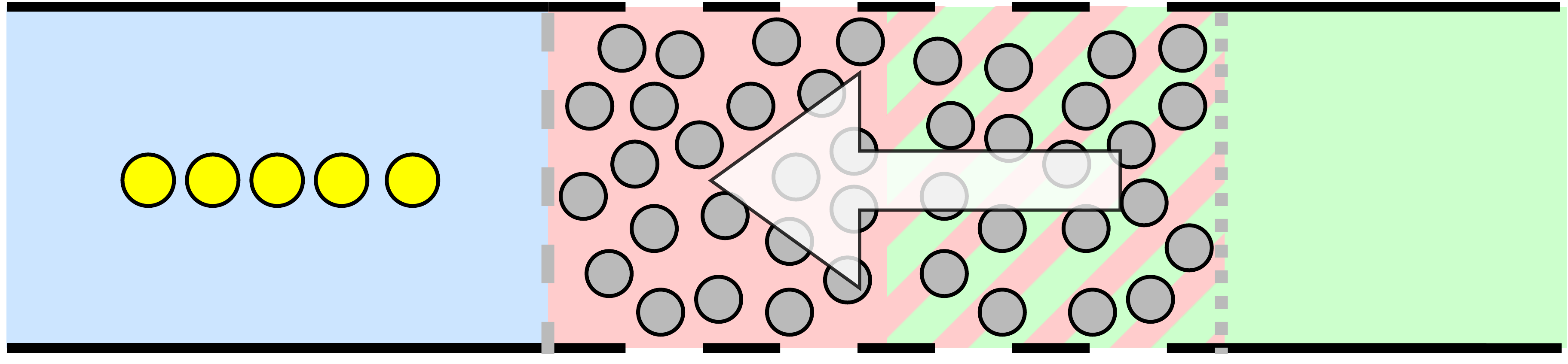}\label{fig:env_scenarios:neg-flow}}
    } \\
    \cline{2-2}
    & \subfloat[Perpendicular-flow crowd]{\includegraphics[width=0.49\linewidth,valign=m]{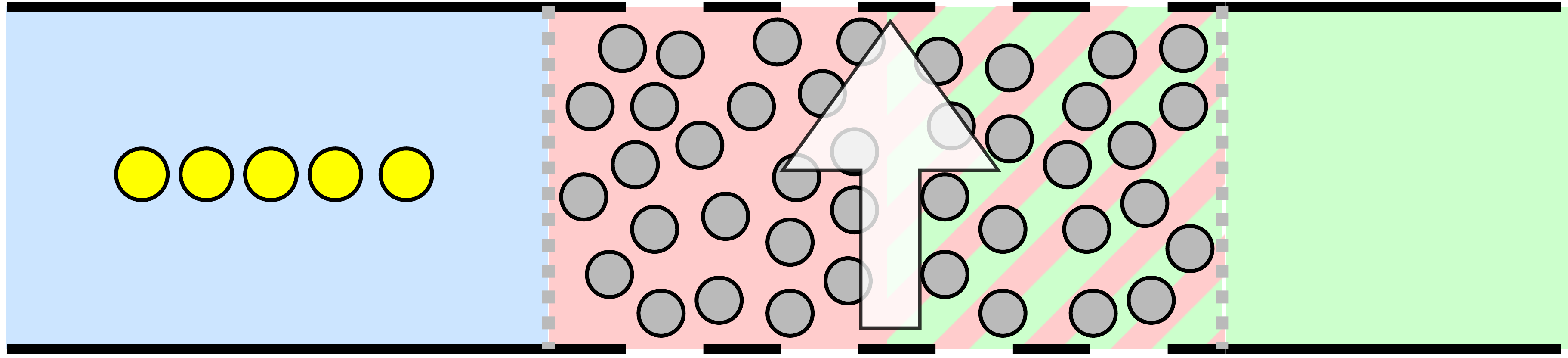}\label{fig:env_scenarios:perpen-flow}} \\
    \end{tabular}
    \caption{Three crowded environments (not to scale). A group of robots (yellow disks) has to move from a start region ($\mathcal{W}_S$) to a goal region ($\mathcal{W}_G$) while traversing a region ($\mathcal{W}_C$) where they encounter a crowd of agents (gray disks). The solid black lines represent the boundary of the environment, while the dashed black lines represent a periodic boundary. (a) Crowd agents seek to remain stationary. (b--c) Crowd agents seek to move in the direction indicated by the white arrow. The dotted gray line further restricts the movements of crowd agents. 
    In (b), any crowd agent crossing the dashed gray line into $\mathcal{W}_S$ is removed and a new crowd agent is added into $\mathcal{W}_C \cap \mathcal{W}_G$.}
    \label{fig:env_scenarios}
\end{figure}

The environment is denoted as a region  $\mathcal{W}\subset\mathbb{R}^2$ (see Fig.~\ref{fig:env_scenarios}) which
consists of:  
    a \textit{start region}, $\mathcal{W}_S$,
    a \textit{crowd region}, $\mathcal{W}_C$, and
    a \textit{goal region}, $\mathcal{W}_G$.


We assume that the closures of the start and goal regions are disjoint (i.e., $\mathring{\mathcal{W}}_S\cap\mathring{\mathcal{W}}_G=\emptyset$) whereas the boundaries of the start and crowd regions intersect (i.e., $\partial\mathcal{W}_S\cap\partial\mathcal{W}_C\ne\emptyset$). The interiors of regions $\mathcal{W}_C$ and $\mathcal{W}_G$ intersect (i.e., $\mathcal{\mathring{W}}_C \cap \mathcal{\mathring{W}}_G \ne\emptyset$).
The intersection between the boundary of $\mathcal{W}$ and the closure of $\mathcal{W}_C$ is composed of two disjoint sets that are associated via a bijective correspondence. The sets are considered equal, an assumption that in our context represents a periodic boundary condition. 
 
The environment contains $n_r$ robots which are modeled as disks of diameter $d_r$. We use $\mathbf{p}^r_i[k] = \left[x^r_i \quad y^r_i \right]^\top$ to denote robot $i$'s position at time $k$. In this instant, the robot occupies region 
$\mathcal{R}_i[k]=\mathcal{B}(\mathbf{p}_i^r[k],\frac{d_r}{2})$, where $\mathcal{B}(\mathbf{p},\epsilon)=\{\mathbf{q}\in\mathcal{W}\mid||\mathbf{p}-\mathbf{q} ||<\epsilon\}$. All robots are initially placed in the start region (i.e., $\forall i: \mathcal{R}_i[0] \subset \mathcal{W}_S$). 

The environment contains $n_c$ crowd agents, each of which has a comfort zone modeled as a disk of diameter $d_c$. We use $\mathbf{p}^c_i[k] = \left[x^c_i \quad y^c_i \right]^\top$ to denote  crowd agent $i$'s position at time $k$. In this instant, its comfort zone is defined by region
$\mathcal{C}_i[k]=\mathcal{B}(\mathbf{p}_i^c[k],\frac{d_c}{2})$. The crowd agents are initially placed in $\mathcal{W}_C$ (i.e., $\forall i: \mathcal{C}_i[0] \subset \mathcal{W}_C$).
For simplicity of notation, we omit variable $k$ when it is clear from the context.
%

The robots' objective is to reach the goal region ($\mathcal{W}_G$). 
At the precise instant the entirety of robot $i$ is contained within the goal region, it is declared to have arrived at the goal: that is, if $\mathcal{R}_i[k^*]\subset\mathcal{W}_G$  but $\mathcal{R}_i[k^*-1]\not\subset\mathcal{W}_G$, then robot $i$ is considered to have reached the goal at time $k^*$. The less time it takes for all robots to have reached the goal, the better.

To reach the goal region ($\mathcal{W}_G$), robot $i$ must traverse parts of the crowd region ($\mathcal{W}_C\setminus \mathcal{W}_G$). At all times, it has to prevent overlapping with the boundary of the environment ($\mathcal{R}_i \cap \partial\mathcal{W} = \emptyset$), other robots ($\forall j: \mathcal{R}_i \cap \mathcal{R}_j=\emptyset$), and the comfort zones of crowd agents ($\forall j: \mathcal{R}_i \cap \mathcal{C}_j=\emptyset$). 
If overlaps occur between a robot and another agent, they are resolved by moving each of the overlapping agents away from each other by half of the overlapped distance. If overlaps occur between an agent (including robots) and a wall, the agent is moved away by the overlapped distance.
Crowd agents are allowed to overlap with other crowd agents. Their interactions with each other is fully dictated by the social force model~\cite{helbing1995}.

We consider three variants of the problem:
\begin{enumerate}
    \item \textit{passive} crowd (see Fig.~\ref{fig:env_scenarios:no-flow}):  Each crowd agent seeks to remain stationary while being repelled by nearby
robots and other crowd agents.
    \item \textit{counter-flow} crowd (see Fig.~\ref{fig:env_scenarios:neg-flow}): Each crowd agent seeks to move toward the start region ($\mathcal{W}_S$) while being repelled by nearby
robots or other crowd agents.
Once fully within the start region, the agent gets removed from the environment, and a new one is inserted in region $
\mathcal{W}_C\cap\mathcal{W}_G$. 
    \item \textit{perpendicular-flow} crowd (see Fig.~\ref{fig:env_scenarios:perpen-flow}): Each crowd agent seeks to move toward the ``upper'' boundary of the crowd region while being repelled by nearby
robots or other crowd agents.
\end{enumerate}
For all variants, the environment contains a periodic boundary, causing crowd agents that leave via the ``upper'' boundary of the crowd region to reenter via the  ``lower'' boundary and vice versa.

In the following, the robots and crowd agents are described in detail.

\subsection{Robot Model}

Robot $i$ has state vector $\mathbf{p}^r_i$ and control inputs $\mathbf{u}^r_i$ given by
\begin{equation*}
\mathbf{p}^r_i = \begin{bmatrix}
x^r_i \quad y^r_i
\end{bmatrix}^\top , \quad
\mathbf{u}^r_i = \begin{bmatrix}
v^r_{x,i} \quad v^r_{y,i}
\end{bmatrix}^\top
\label{eq:wq}
\end{equation*}
where the inputs $v^r_{x,i}$ and $v^r_{y,i}$ are its linear velocity in $x$ and $y$, respectively. 

The robot dynamics correspond to a holonomic robot. The state of robot $i$ at time $k+1$ is given by
\begin{equation*}
\mathbf{p}^r_{i}[k+1]  = \mathbf{p}^r_{i}[k]  + 
\mathbf{u}^r_i[k]
\Delta t
    \label{eq:rhsmodel}
\end{equation*}
where $\mathbf{u}^r_i[k]$ is the control input at time $k$
and $\Delta t$ is the sampling time. The robot's control input (i.e. velocity) is bounded, 
$||\mathbf{u}^r_i[k]||\le v^{r,max}$.
Although the definition of control inputs assumes the robots to share a common orientation, this is merely to simplify notation.


We assume each robot has a unique ID. Robot $i$ can interact with other robots, agents and objects within sensing range, $r^r_s$. It detects (i) the IDs and relative positions of the other robots, $\mathcal{N}^r_i$ and $\mathbf{r}^r_{i,j}$, (ii) the positions of the crowd agents relative to itself, $\mathcal{Q}^c_i$, (iii) and a closest point of the environment boundary relative to itself, $\mathbf{q}^{\partial\mathcal{W},min}_i$, if any.
Formally, $\mathcal{N}^r_i=\{ j\in\{1,\dots,i-1, i+1, \dots, n_r\} \mid || \mathbf{p}^r_j - \mathbf{p}^r_i || < r^r_s \}$,  $\forall j\in\mathcal{N}^r_i, \mathbf{r}^r_{i,j}=\mathbf{p}^r_j - \mathbf{p}^r_i$,
$\mathcal{Q}^c_i=\{ (\mathbf{p}^c_j - \mathbf{p}^r_i) \mid j\in\{1,\dots,n_c\}: || \mathbf{p}^c_j - \mathbf{p}^r_i || < r^r_s \}$, and $\mathbf{q}^{\partial\mathcal{W},min}_i =\{ (\mathbf{q} - \mathbf{p}^r_i) \mid \argmin_{\mathbf{q}\in\partial\mathcal{W}\cap\mathcal{B}(\mathbf{p}^r_i,r^r_s)} || \mathbf{q} - \mathbf{p}^r_i || \}$.\footnote[1]{Due to the structure of the environment, at most a single $\mathbf{q}$ minimizes the expression.} From this, the robot obtains $\mathcal{Q}^r_i=\{ \mathbf{r}^r_{i,j} \mid j\in \mathcal{N}^r_i\}$. Robot $i$ can also communicate with other robots in its neighborhood, $\mathcal{N}^r_i$.

\subsection{Crowd Agent Model}

The crowd agents follow the social crowd model as described in~\cite{helbing1995}, widely used for simulating pedestrian crowds. They can move in all directions. They seek to move in a reference direction while being repelled by other agents, robots, and walls. Each crowd agent is provided with (i) a vector of the desired velocity, (ii) the position and velocity of nearby crowd agents, and (iii) the position of the closest point of the environment's boundary. Note that the social crowd model considers interactions within a homogeneous group of agents---pedestrians. To model interactions between crowd agents and robots, we assume that crowd agents are repelled equally strongly by robots and their peers.


\section{Control Strategies}
\label{CS}
In this section, we present the control strategies for the robots. All strategies are based on the distributed artificial potential field (APF) controller, originally presented in~\cite{howard2002mobile}. 
In our work, it causes the robot to be attracted toward the goal region while being repelled by the crowd agents, other robots, and the environment boundary.

The artificial force experienced by robot $i$ is given by
\begin{equation}
\begin{aligned}
\mathbf{F}^r_i ={} 
& k^r_{g} \frac{\mathbf{g}^r_i}{||\mathbf{g}^r_i||} 
- k^r_{r} \sum_{\mathbf{q} \in \mathcal{Q}^r_i }\frac{\mathbf{q}}{\left(||\mathbf{q}||\right)^3}
- k^r_{c} \sum_{\mathbf{q} \in \mathcal{Q}^c_i} \frac{\mathbf{q}}{\left(||\mathbf{q}||\right)^3} 
 - k^r_{b}  \frac{\mathbf{q}^{\partial\mathcal{W},min}_i}{\left(||\mathbf{q}^{\partial\mathcal{W},min}_i||\right)^3}
\end{aligned}
\label{eq:apf_forces}
\end{equation}
where $k^r_{g}$, $k^r_{r}$, $k^r_{c}$, $k^r_{b}\in \mathbb{R}^+$ are weights, $\mathbf{g}^r_i$ serves as an individual goal position for robot $i$ and is defined with respect to its local reference frame.

The desired inputs, $\mathbf{u}^{r,des}_i := \begin{bmatrix}
v^{r,des}_{x,i} & v^{r,des}_{y,i}
\end{bmatrix}^\top$, are derived from the force vector as $\mathbf{u}^{r,des}_i= \mathbf{F}^r_i  \Delta t.$ To respect the actuator limits, the desired linear velocities may need to be modified. We therefore obtain the velocities that the motors need to realize as $\mathbf{u}^r_i = \mathbf{u}^{r,des}_{i} \cdot \mathcal{G}(\mathbf{u}^{r,des}_{i})$, where
        $\mathcal{G}(\mathbf{v})=1$ 
 if $||\mathbf{v}|| \leq v^{r,max}$, 
and $v^{r,max}/ ||\mathbf{v}||$, otherwise.

Using the APF architecture, three control strategies are considered:
\begin{enumerate}
    \item \textit{Platoon}: 
    Robot 1 uses as an individual goal position, $\mathbf{g}^r_1$, a landmark from within the goal region.
    Robot $i \in \{2,\dots,n_r \}$ uses as individual goal position,  $\mathbf{g}^r_i = \mathbf{r}^r_{i,i-1}(1-\frac{d_r}{||\mathbf{r}^r_{i,i-1}||})$.
Moreover, if the distance to its follower, $||\mathbf{r}^r_{i+1}||$, exceeds threshold value $r^r_p < r^r_s$, 
robot $i \in \{1,\dots,n_r-1 \}$ pauses, by setting the desired inputs to $\mathbf{u}_i^{r,des}=\mathbf{0}$.
    \item \textit{Greedy}: Robot $i$ uses as individual goal position, $\mathbf{g}^r_i$, a landmark from within the goal region.    
    %
    \item \textit{Adaptive Platoon}: Algorithm \ref{algorithm} details the strategy. Initially, robot 1 has no leader (i.e., $L=\text{nil}$), whereas robot $i>1$ uses robot $L=i-1$ as its leader.    
  Robot $i$ removes its association with leader $L$ if the latter is outside the sensing range, occluded by the crowd (i.e., the line segment between the robots intersects with a crowd agent), or deviates from the direction of robot $i$'s individual landmark within the goal region by at least angle $\alpha$.
  Robot $i$ becomes associated with robot $L=j$ if the latter has no follower yet, is no further than $r_{\beta}< r^r_s$ away, and does not deviate from the direction of robot $i$'s individual landmark within the goal region by more than angle $\beta$.
\end{enumerate}

\begin{algorithm}[t]
\caption{Adaptive Platoon}
\begin{algorithmic}[1]
\Procedure{AdaptivePlatoon}{}
    \State $\mathbf{g}^r_i=$ \Call{ObtainGoalRegionLandMark}{} \Comment{greedy strategy}
    \If{$L\ne \text{nil}$}
        \If{$L\not\in \mathcal{N}^r_i$ \textbf{or} 
            \Call{IsOccluded}{$L$} \textbf{or} 
            {$\sphericalangle (\mathbf{r}^r_{i,L}, \mathbf{g}^r_i)\ge \alpha$}}
            \State $L=\text{nil}$
        \EndIf
    \Else
        \For{\textbf{each} $j$ \textbf{in} $\mathcal{N}^r_i$}
            \If{\Call{NotFollowed}{$j$} \textbf{and}  $||\mathbf{r}^r_{i,j}||< r_\beta$ \textbf{and} 
                $\sphericalangle (\mathbf{r}^r_{i,j}, \mathbf{g}^r_i)\le \beta$}
                \State $L=j$
                \State \textbf{break}
            \EndIf
        \EndFor
    \EndIf
    \If{$L\ne \text{nil}$}
        \State $\mathbf{g}^r_i = \mathbf{r}^r_{i,L}(1-\frac{d_r}{||\mathbf{r}^r_{i,L}||})$\Comment{platooning strategy without pausing}
        \State notify $L$ it is being followed
    \EndIf
    \State \Call{APF}{$\mathbf{g}^r_i$}  \Comment{call artificial potential field controller}
\EndProcedure
\end{algorithmic}
\label{algorithm}
\end{algorithm}


\section{Simulation Setup}
\label{SS}

\begin{table}[b]
    \centering
    \caption{Parameters used for the robot hardware and control strategy.}
    \begingroup
    \renewcommand{\arraystretch}{1} 
    \begin{tabular}{cccccc}
        \toprule
        \textbf{Parameter} & \textbf{Value} & \textbf{Parameter} & \textbf{Value} & \textbf{Parameter} & \textbf{Value}\\
        \cmidrule(lr){1-2}
        \cmidrule(lr){3-4}
        \cmidrule(lr){5-6}
        $d_r$ (m) & 0.3 & $k^r_g$ & 3.5  & $r^r_p$ (m) & 0.6\\
        $v^{r,max}$ (ms$^{-1}$) & 0.6 & $k^r_{r}$ & 0.2 & $r_{\beta}$ (m) & 0.6\\
        $r^r_s$ (m) & 1.5 & $k^r_{c}$ & 0.1  & $\alpha$ (\textdegree) & 20\\
        & &  $k^r_{b}$ & 0.1 & $\beta$ (\textdegree) & 5\\
        \bottomrule
    \end{tabular}
    \endgroup
    \label{tab:design-params}
\end{table}

We use a numerical simulator with a sampling time of $\Delta t=0.1$\,s. The simulations are performed using an Intel Xeon Platinum 8358 CPU core with 10\,GB of memory in a High-Performance Computing cluster that uses Python $3.8.18$.

Regions $\mathcal{W}_S$, $\mathcal{W}_C$, $\mathcal{W}_G$ are rectangular and of dimensions (H\,$\times$\,W)  5\,m$\times  \infty$\,m,  5\,m$\times$10\,m, and  5\,m$\times  \infty$\,m. The left side of $\mathcal{W}_G$ starts at the horizontal center of $\mathcal{W}_C$,  as shown in Fig.~\ref{fig:env_scenarios}.

Table~\ref{tab:design-params} lists the robot-specific parameters used for the computational experiments.
Each robot $i$ experiences a zero mean Gaussian noise in their inputs $v^r_{x,i}$ and $v^r_{y,i}$ with standard deviations $\sigma_{v_x} = 0.05 \cdot v^r_{x,i}$ and $\sigma_{v_y} = 0.05 \cdot v^r_{y,i}$, respectively. 
Each crowd agent has a comfort zone of diameter $d_c=0.3$\,m. Its desired direction of motion ($\mathbf{e}^c_i$, see~\cite{helbing1995}) is $\mathbf{e}^c_i = \mathbf{0}$, $\mathbf{e}^c_i = [-1 \quad 0]^\top$, or $\mathbf{e}^c_i = [0 \quad 1]^\top$ for the passive, counter-flow and perpendicular-flow scenarios, respectively.

For all strategies, a group of $n_r = 10$ robots is initialized in a \emph{line} formation that has a spacing between robots of 0.5\,m, and is centered vertically.  For the greedy strategy, in addition, a \emph{random} formation is tested:
The robots are placed with the same spacing of 0.5\,m in the $x$-axis but the $y$ coordinate is uniformly randomly sampled from the range of coordinates that are at least 0.5\,m away from the ``lower'' and ``upper'' boundary.
For either formation, the center point of the robot closest to the crowd region is 3\,m away from the crowd region. At all times, robot $i$ chooses as landmark $\mathbf{g}^r_i$ the point on the (infinitely far) right-hand side of the rectangular goal region that is closest to the robot at time 0. In other words, the corresponding unit vector $\frac{\mathbf{g}^r_i}{||\mathbf{g}^r_i||}$ in Equation \ref{eq:apf_forces} becomes $[1 \quad 0]^\top$.

For the sake of simplicity, leader-follower pairs are pre-specified for the platooning strategies. The robot closest to the goal region is assigned an ID of 1 and subsequent IDs are assigned from right to left. 
Given the regular nature of the formation, the IDs could be automatically assigned as well.

The desired number of crowd agents, $n_c$, depends on the chosen density $\rho$ for a given simulation trial,
$n_{c} = \left\lfloor\frac{\rho \cdot A}{ \frac{\pi}{4} \cdot {d_c}^2}\right\rfloor$,
where $A$ is the area of region $\mathcal{W}_C$ (see Fig.~\ref{fig:env_scenarios}).
Positions for crowd agents are chosen at random using a uniform distribution. Where the sampled position of a crowd agent would cause it to overlap with a previously placed crowd agent, the sampling process is repeated (for a maximum of 10000 times).

\section{Insights from Using the Social Force Model}
\label{MI}

Our crowd model is based on the original social force model~\cite{helbing1995}. The latter
considers crowd agents as volume-less particles. Consequently, it is possible for two crowd agents to be, or move, arbitrarily close in proximity to each other.
Although our model includes a comfort zone for each crowd agent, the interactions among crowd agents are fully governed by the social force model~\cite{helbing1995}. Hence, it is common for their comfort zones to overlap. As a crowd agent does not repel more strongly from robots than other crowd agents, its comfort zone can overlap with a robot's body too. This can be observed even where robots are entirely static, especially when located in the crowd's principal path of motion. 
In this study, we quantify the disruption caused by the robots to the crowd, measuring the number of times a robot's body intersects with the comfort zone of a crowd agent. Note that our setting is conservative. If the crowd agents repelled more strongly from robots than from each other, one would expect to observe fewer interceptions.

\begin{figure}[t]
    \centering
    \includegraphics[width=0.825\linewidth]{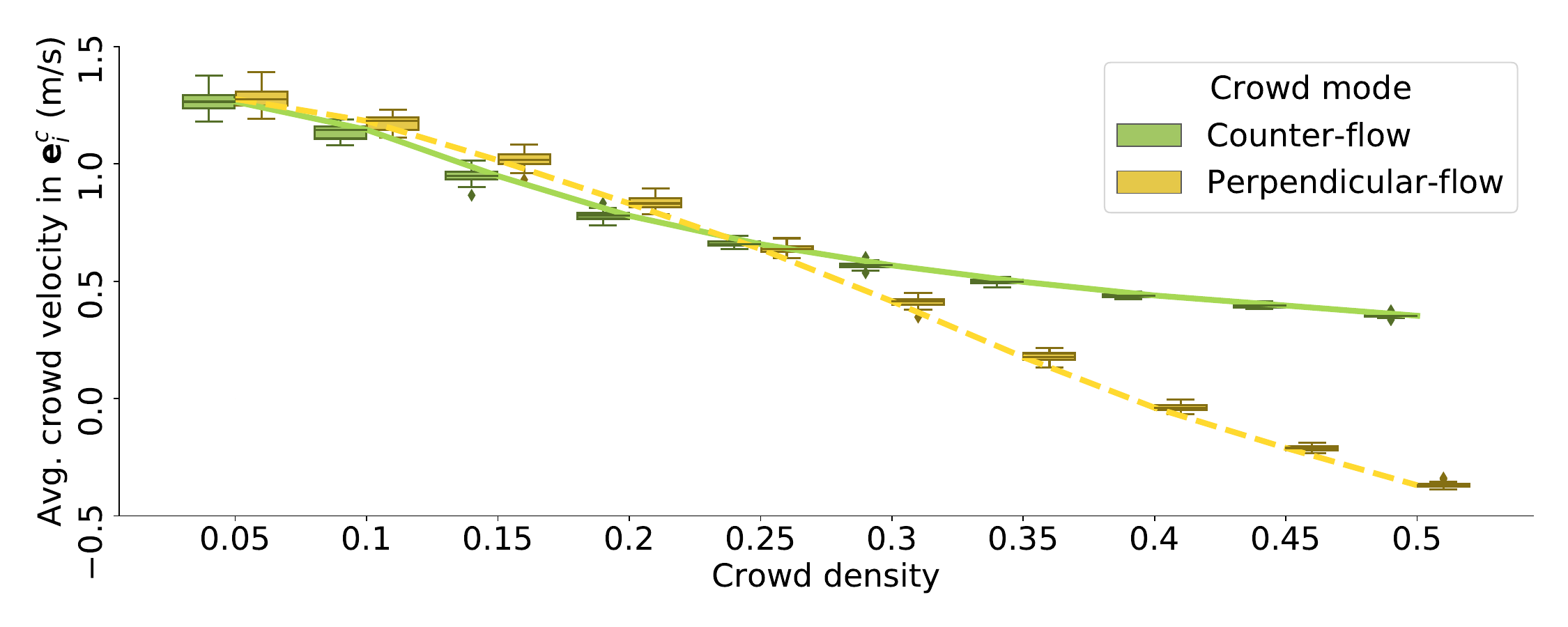} 
    \caption{Average crowd velocity in their desired direction. The solid green and dashed yellow lines show how the average crowd velocity decreases as the crowd densities for counter-flow and perpendicular-flow increase, respectively.}
    \label{fig:lessons}
\end{figure}

To better understand the social force model~\cite{helbing1995}, we first tested the crowds in isolation, that is, without any robot present.
We observed that the average velocity of the crowd agents decreases with crowd density, and that the crowds may even reverse their direction of movement (see Fig.~\ref{fig:lessons}, 30 simulation trials per setting).
This can be attributed to the non-isotropic
repulsion field~\cite{helbing1995}, 
where the focal agent is repelled more strongly by agents in its front (i.e., its desired direction) than those behind.
When the direction of motion is toward the periodic boundary, such as in the perpendicular-flow crowd scenario, the average velocity of the crowd reduces about linearly, with the flow even inverting its direction at a density of around 0.4. In contrast, in the counter-flow crowd scenario, the velocity reduces asymptotically to a fraction of the nominal velocity. In this case, we observe non-uniform crowd densities, with low densities towards the crowd agents' destination boundary and jammed agents towards the opposite boundary. 
Due to the flow reduction (and even inversion), we decided to limit the rest of the study to crowd densities up to 0.3. Densities higher than 0.3 will be considered in future work, however, they may require bespoke crowd models.

\section{Results and Discussion}
\label{RD}

We evaluate the control strategies by conducting tests across crowd densities ranging between 0 to 0.3.
Each configuration is tested for 30 trials. If any robot fails to reach the goal region before a timeout of 900\,s, the simulation terminates. We evaluate the performance using the following criteria:
\begin{itemize}
\item \textit{Time to goal}: 
We report the time from the first robot entering the crowded region to the last robot reaching the goal region. Hence, the time required for the first robot to reach the crowded region is discarded.
\item \textit{Comfort zone interception}:  
We report the cumulative number of interceptions at any time between any robot-crowd agent pair. Such interceptions are considered a disruption toward the crowd agents.
\end{itemize}

The results for all variants are presented in the following.
Video recordings can be found at~\cite{argote-gerald_2024_supplementary}.
The simulation source code is available at \cite{argote-gerald_2024_source}.

\subsection{Passive Crowd Scenario}


\begin{figure}[t]
    \centering
    \begingroup
    \setlength{\tabcolsep}{0pt} 
    \renewcommand{\arraystretch}{1} 
    \begin{tabular}{cc}
    \multicolumn{2}{c}{
    \includegraphics[width=0.99\linewidth]{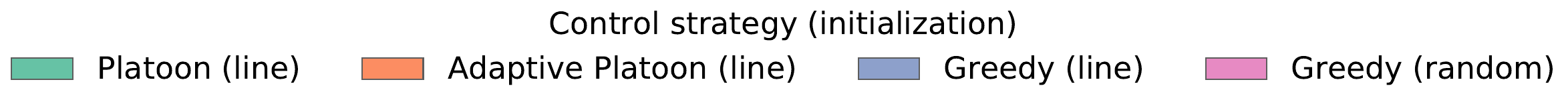}} \\
    \includegraphics[width=0.49\linewidth,valign=m]{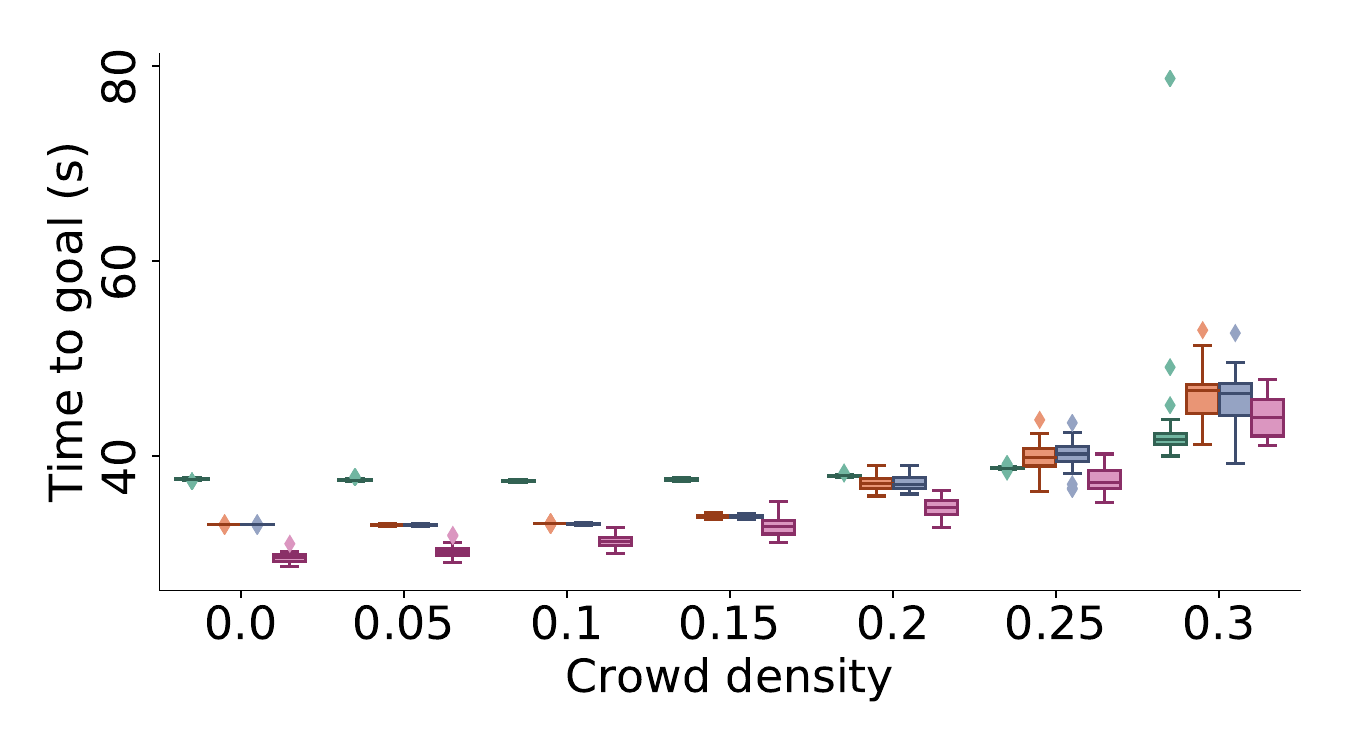} \speciallabel{\ref{fig:passiveresults}a}{fig:passiveflowtime}   &
    \includegraphics[width=0.49\linewidth,valign=m]{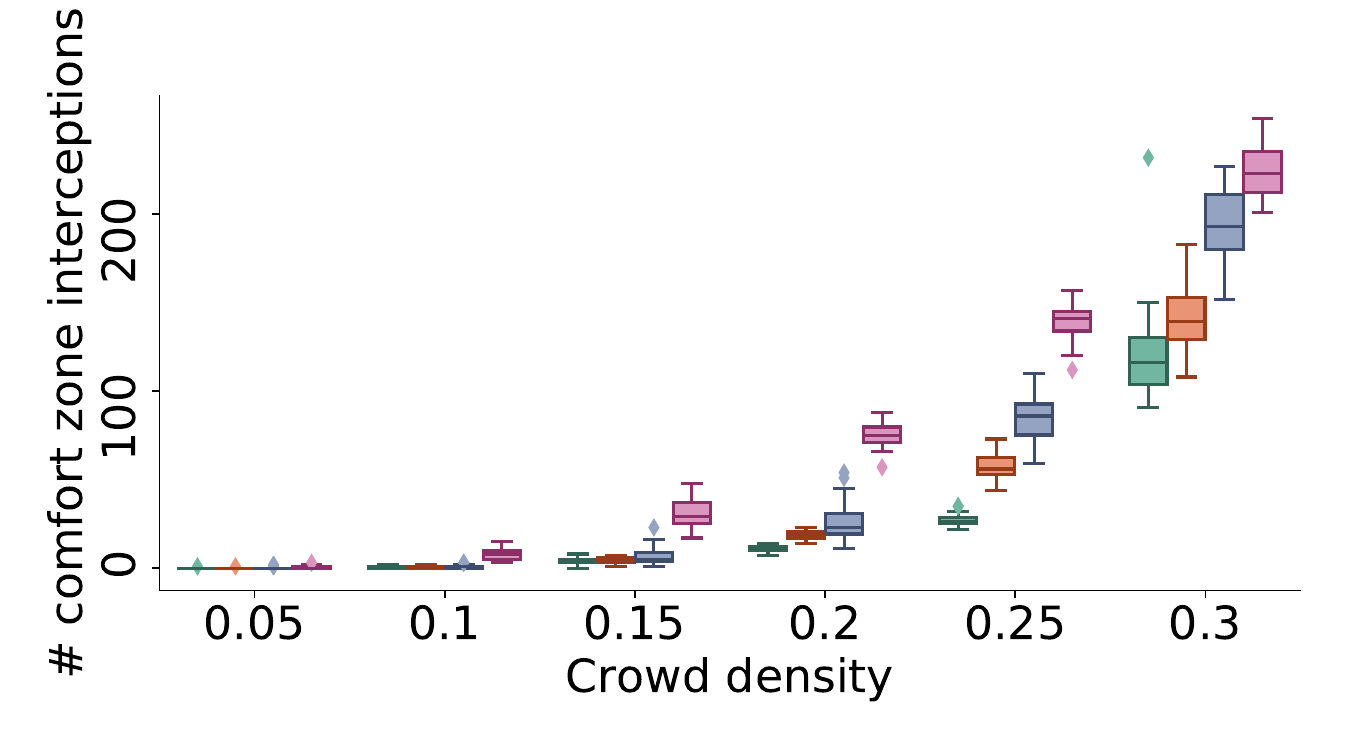} \speciallabel{\ref{fig:passiveresults}b}{fig:passiveflowcol}\\ 
     \small (a) & \small (b)
    \end{tabular}
    \endgroup
    \caption{Passive crowd navigation: (a) time taken for all robots to reach the goal region, and (b) number of comfort zone interceptions with the crowd.}
    \label{fig:passiveresults}
\end{figure}

Fig.~\ref{fig:passiveflowtime} shows that platooning robots take longer to reach the goal when the crowd density is 0.2 or less. In such low-density situations, the greedy strategy seems to perform particularly well, whereas the platooning strategy seems to constrain unnecessarily the movements of individual robots.
When starting from a random configuration, the ``greedy'' robots arrive even faster at the goal, 
as making use of the entire width of the regions to travel, thereby maximizing the flow of robots while minimizing their interference. 
However, at higher crowd densities (i.e., $\ge0.25$), the platoon formation becomes beneficial as all but the first robot can take advantage of the gaps created by the preceding robots.

Fig.~\ref{fig:passiveflowcol} shows that the number of comfort zone interceptions is significantly lower for the platoon strategy than for the greedy strategy (apart from low crowd densities), suggesting that the robots when forming a platoon are less likely to cause disruption to the crowd.
Compared to the reactive strategy, the adaptive platoon strategy causes less disruption to the crowds, and reaches the goal about as fast, presumably owing to robots not having to wait for followers.



\subsection{Counter-flow Crowd Scenario}

\begin{figure}
    \centering
    \begingroup
    \setlength{\tabcolsep}{0pt} 
    \renewcommand{\arraystretch}{1} 
    \begin{tabular}{cc}
    \multicolumn{2}{c}{
    \includegraphics[width=0.99\linewidth]{Figures/legend.pdf}} \\
    \includegraphics[width=0.49\linewidth,valign=m]{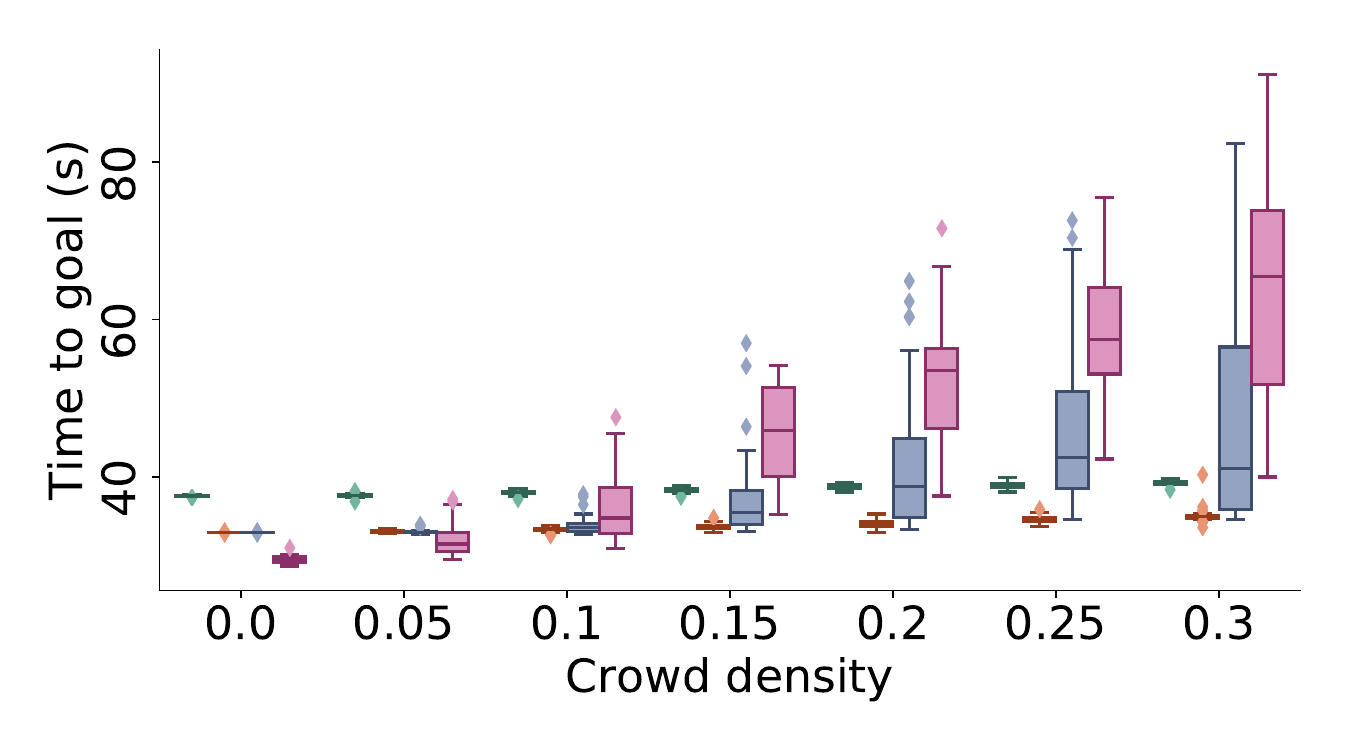}
    \speciallabel{\ref{fig:resultssc2}a}{fig:negflowtime} &
    \includegraphics[width=0.49\linewidth,valign=m]{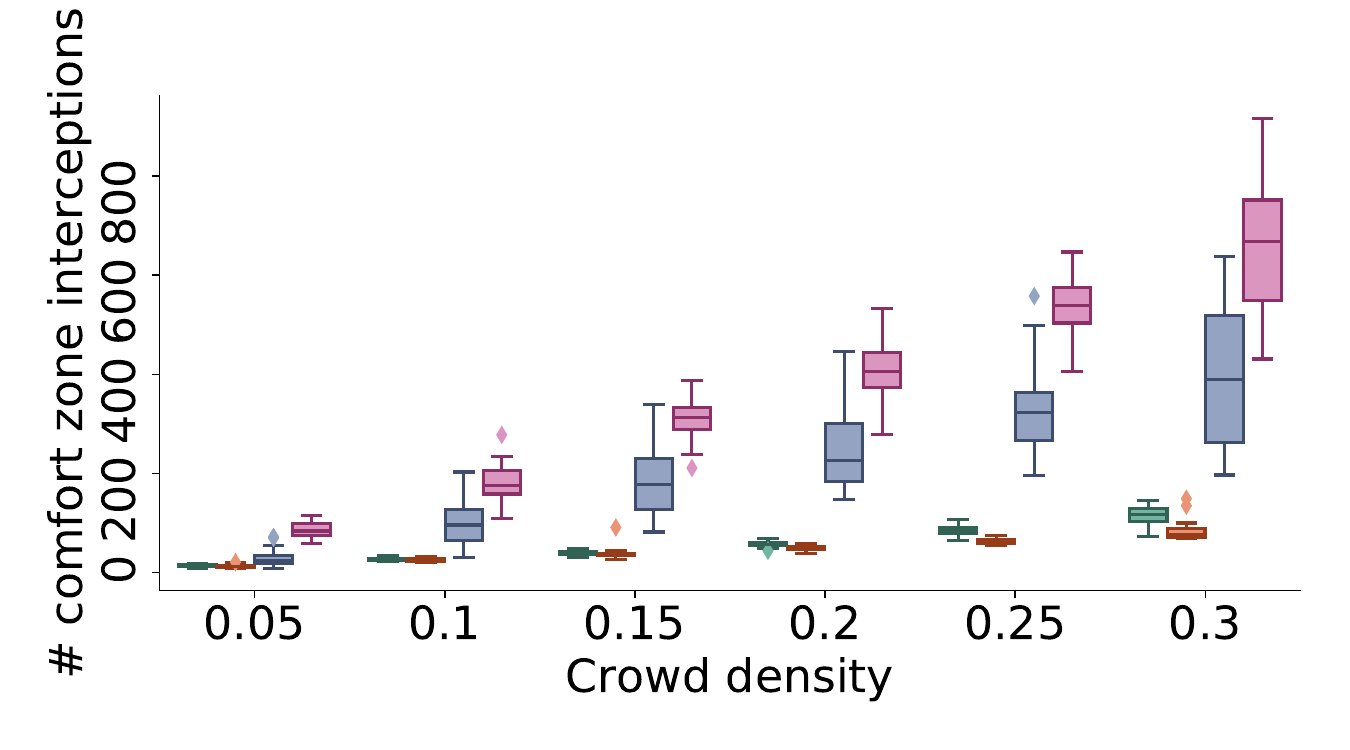} 
    \speciallabel{\ref{fig:resultssc2}b}{fig:negflowcol}\\ 
    \small (a) & \small (b)
    \end{tabular}
    \endgroup
    \caption{Counter-flow crowd navigation: (a) time taken for all robots to reach the goal region, and (b) number of comfort zone interceptions with the crowd.}
    \label{fig:resultssc2}
\end{figure}



The benefit of platoon formation is particularly notable for crowds moving in the opposite direction to the robots.
Fig.~\ref{fig:negflowtime} shows that the platoon strategies outperform the greedy strategy in the time to reach the goal for high crowd densities.
While for the greedy strategy the time to reach the goal tends to increase with crowd density, and greatly varies, for the platoon strategy it remains fairly constant and far more consistent.
Moreover, the robots using the adaptive platoon strategy are consistently faster than those using the basic platoon strategy, suggesting that the added flexibility helps the group to choose a configuration most suited for the particular crowd density.

The number of comfort zone interceptions is significantly less for robots employing the platoon strategies (see Fig.~\ref{fig:negflowcol}),  similar to what was observed for passive crowds. The number of comfort zone interceptions rises steeply for the greedy strategies. This could be attributed to an increased likelihood of the robots, when spread out across the width of the environment, to directly confront crowd agents.


\begin{figure}[tb]
    \centering
    \begingroup
    \setlength{\tabcolsep}{0pt} 
    \renewcommand{\arraystretch}{1} 
    \begin{tabular}{cc}
    \multicolumn{2}{c}{
    \includegraphics[width=0.99\linewidth]{Figures/legend.pdf}} \\
    \includegraphics[width=0.49\linewidth]{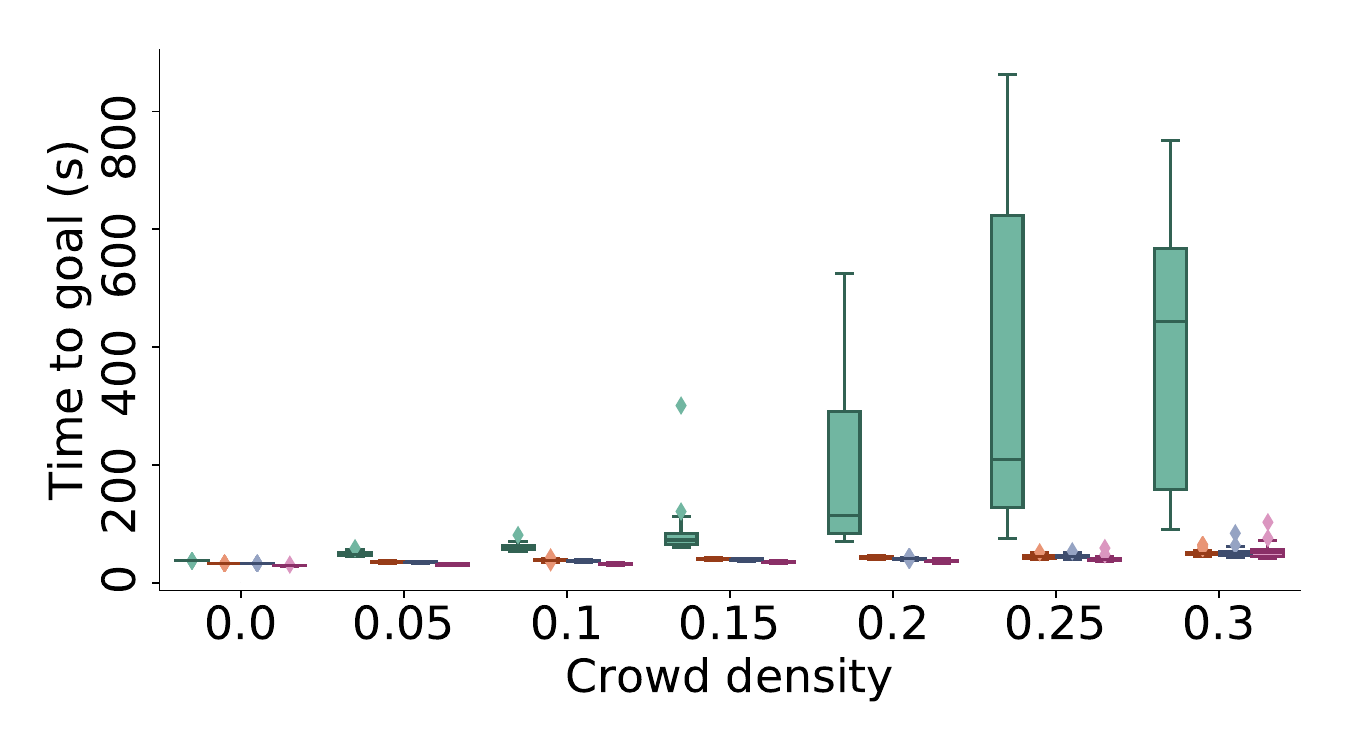} 
    \speciallabel{\ref{fig:resultssc3}a}{fig:perpflowtime} &
    \includegraphics[width=0.49\linewidth]{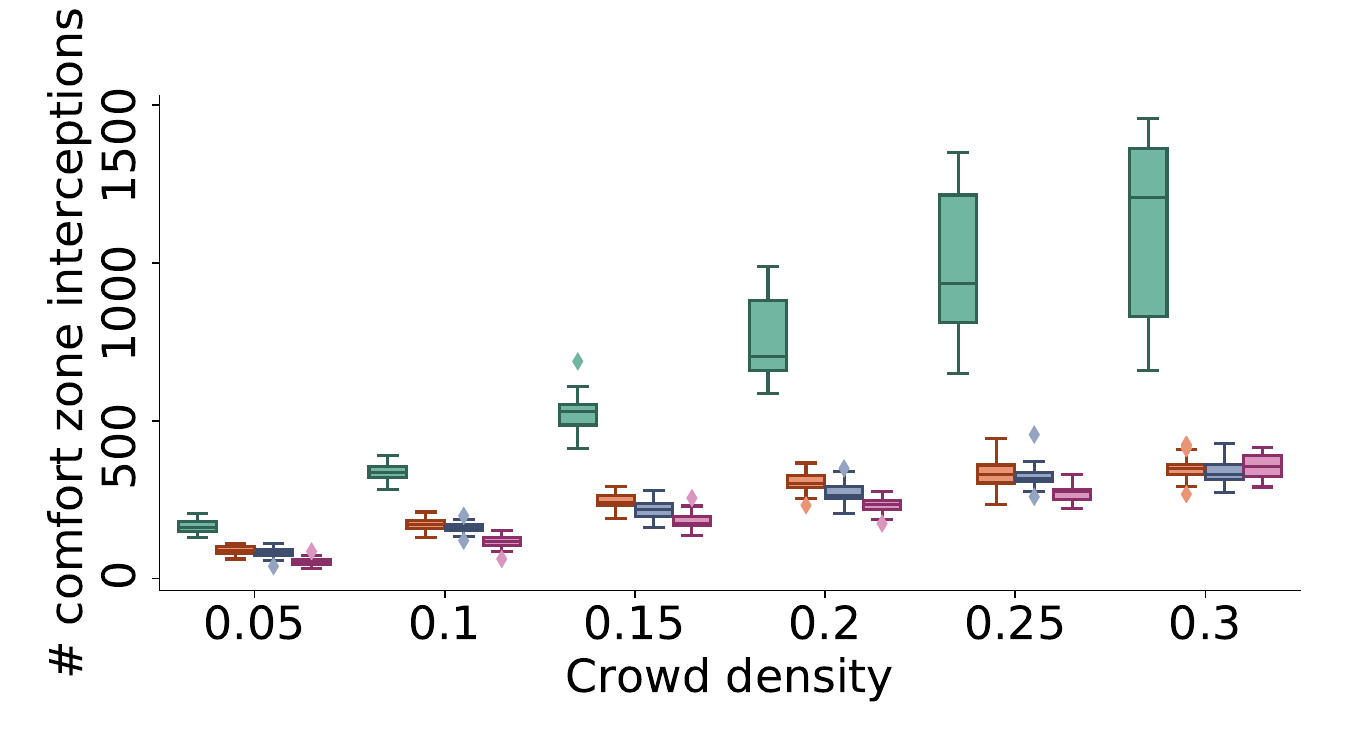} 
    \speciallabel{\ref{fig:resultssc3}b}{fig:perpflowcol}\\ 
    \small (a) & \small (b)
    \end{tabular}
    \endgroup
    \caption{Perpendicular-flow crowd navigation: (a) time taken for all robots to reach the goal region, and (b) number of comfort zone interceptions with the crowd. The platoon (green) failed to reach the goal before the timeout of 900\,s for densities of 0.2, 0.25, and 0.3 in 20\%, 33\%, and 67\% of the trials, respectively.}
    \label{fig:resultssc3}
\end{figure}

\subsection{Perpendicular-flow Crowd Scenario}

Fig.~\ref{fig:resultssc3}a shows that the greedy strategy achieves the best time-to-goal performance for all crowd densities. This can be attributed to the perpendicular flow, which increases the likelihood of crowd agents obstructing members in the middle of the platoon formation, thereby causing their respective leader agents to pause. The difference in time-to-goal performance between the platoon and greedy strategies gets amplified as the density increases. 
The adaptive platoon strategy performs on par with the greedy strategy.
The robots in the adaptive platoon split when it is difficult or not effective to maintain the platoon formation upon encountering the crowd moving perpendicular to their desired direction.

Fig.~\ref{fig:resultssc3}b reveals the disruptive nature of the platoon strategy in this setting. The adaptive platoon strategy by contrast performs almost as well as the reactive strategy.

\subsection{Scalability}

\begin{figure}[tb]
    \centering
    \begingroup
    \setlength{\tabcolsep}{3pt} 
    \renewcommand{\arraystretch}{1} 
    \begin{tabular}{cc}
    \multicolumn{2}{c}{
    \includegraphics[width=0.99\linewidth]{Figures/legend.pdf}} \\
    \includegraphics[width=0.49\textwidth,valign=m, trim={0 0 0 0}, clip]{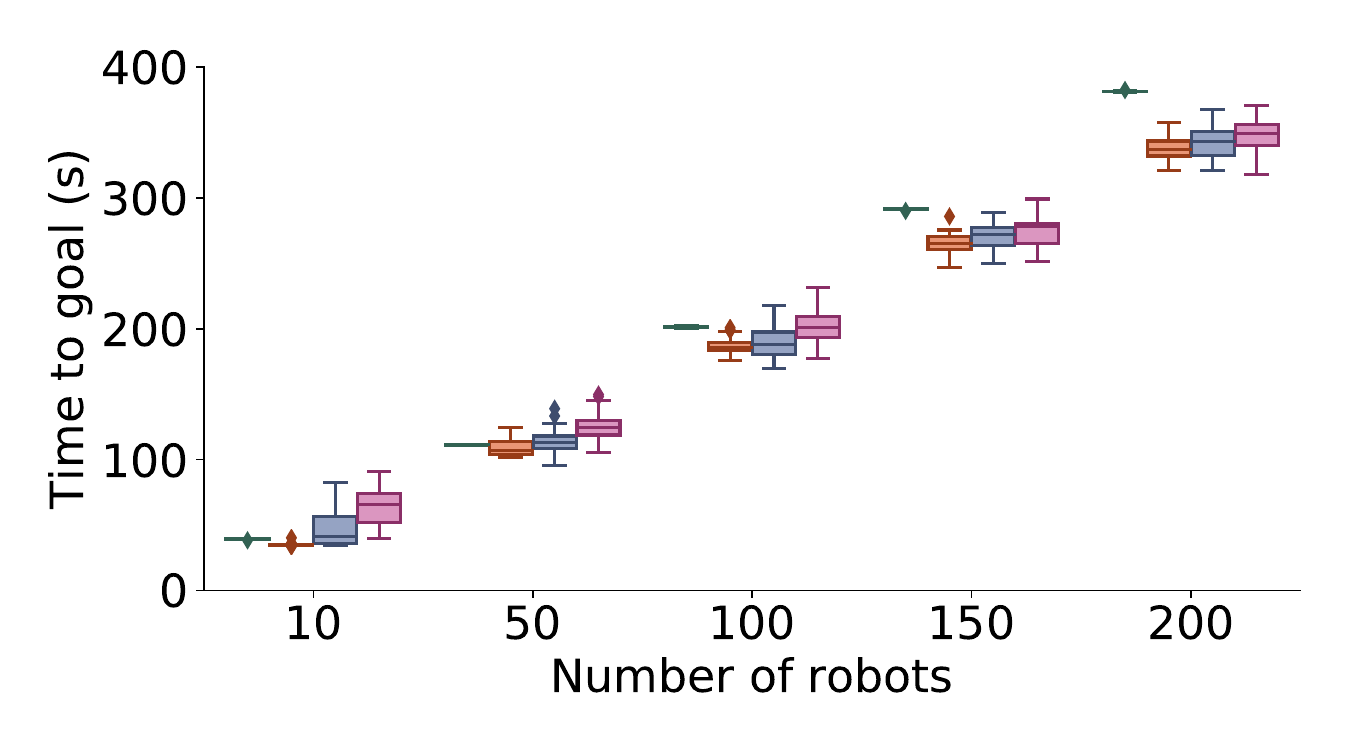} &
    \includegraphics[width=0.49\textwidth,valign=m, trim={0 0 0 0}, clip]{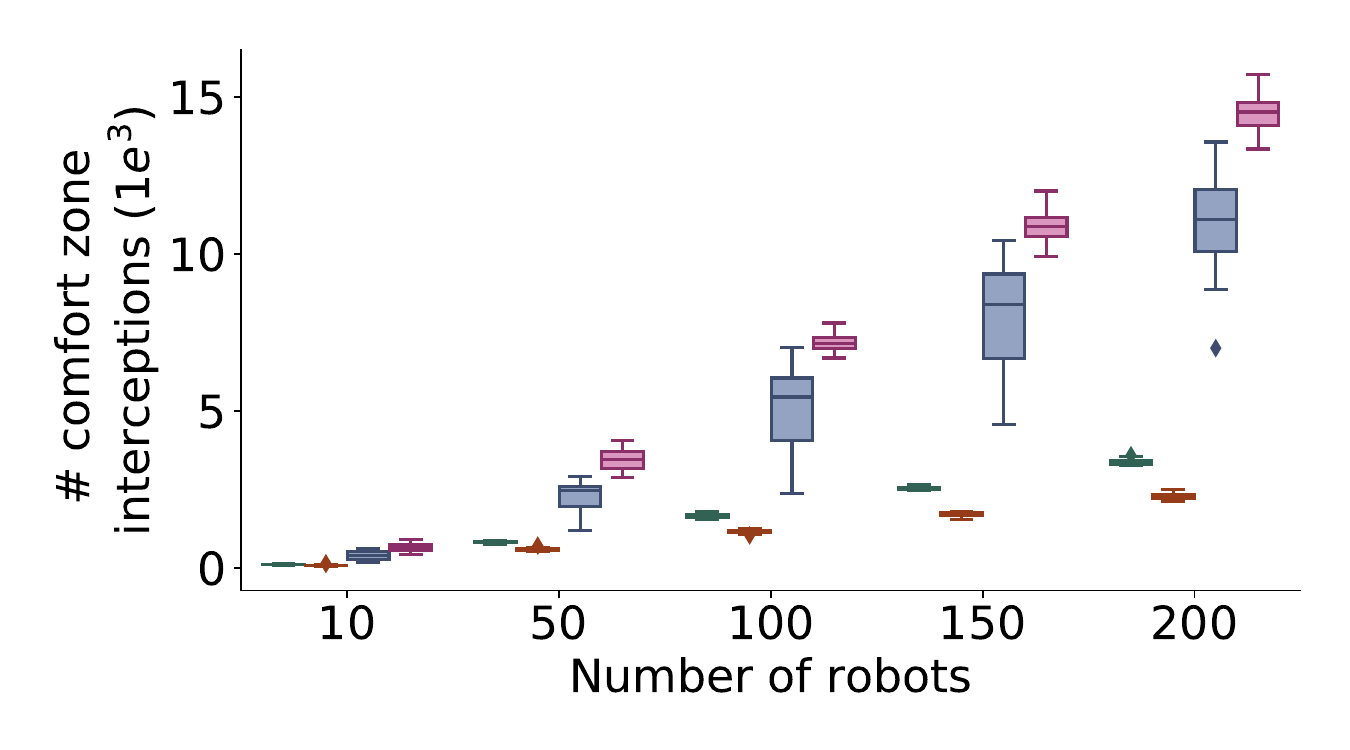} \\
    \small (a) & \small (b)
    \end{tabular}
    \endgroup
    \caption{Scalability analysis for navigating counter-flow crowd of density 0.3: (a) time taken for all robots to reach the goal region, and (b) number of comfort zone interceptions with the crowd.}
    \label{fig:scalability}
\end{figure}

To examine how the number of robots impacts performance, 
we consider counter-flow crowds of density 0.3 and with up to 200 robots. For each setting, 30 trials are performed.
Fig.~\ref{fig:scalability}a shows that for up to 50 robots the platoon formation outperforms the greedy strategy, being slower for a higher number of robots. This could be due to robots pausing to allow followers to catch up (i.e., where the distance constraint imposed by $r^r_p$ is not met), which can lead to compounding effects as the number of robots increases.  
Fig.~\ref{fig:scalability}b shows that irrespective of the robot count, both platoon strategies consistently cause fewer comfort zone interceptions within the crowd compared to the greedy strategy. 
The adaptive platoon outperforms the other strategies for either performance metric for all numbers of robots. This suggests that maintaining ``highways''---achieved through a less restrictive platoon formation---is an effective strategy for navigating through counter-flow crowds.

\begin{figure}[t]
    \centering
    \begingroup
    \setlength{\tabcolsep}{1pt} 
    \renewcommand{\arraystretch}{1} 
    \begin{tabular}{cccccc}
    \includegraphics[width=0.16\linewidth]{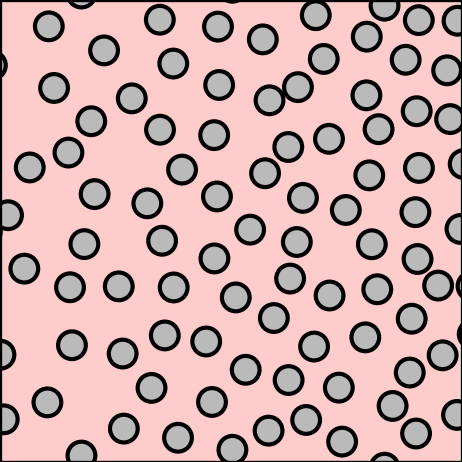} & 
    \includegraphics[width=0.16\linewidth]{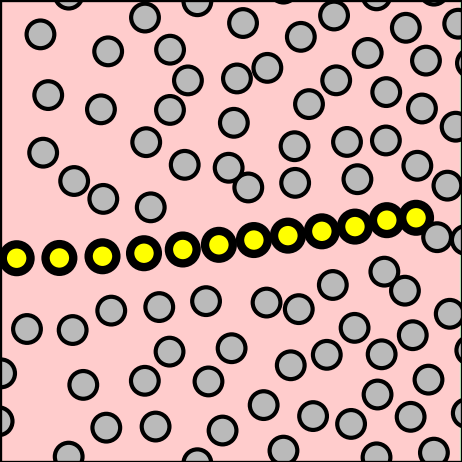} &
    \includegraphics[width=0.16\linewidth]{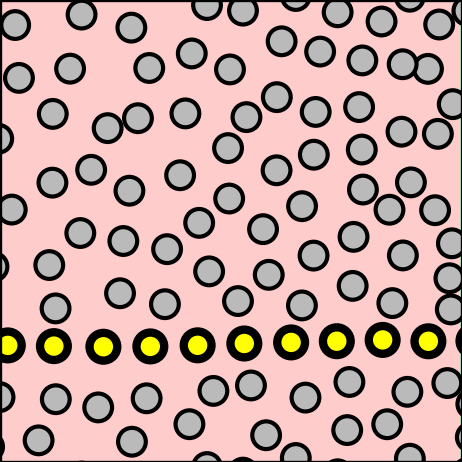} &
    \includegraphics[width=0.16\linewidth]{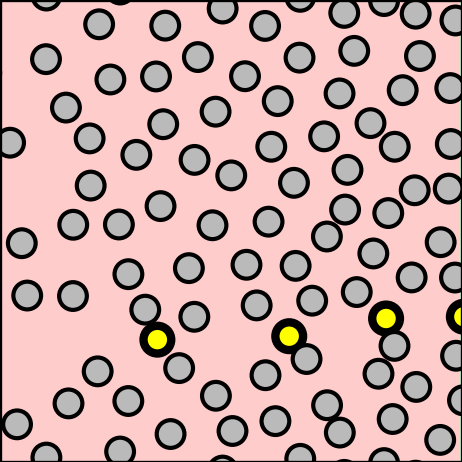} &
    \includegraphics[width=0.16\linewidth]{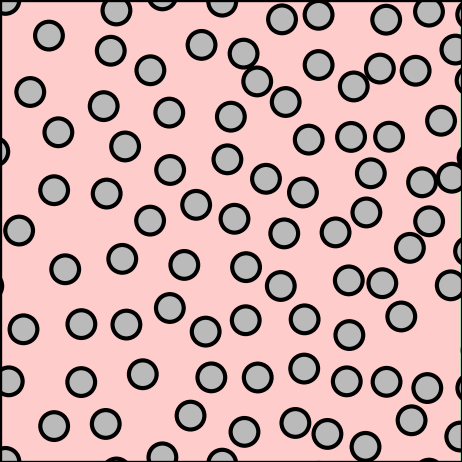} &
     \\ 
    
    \includegraphics[width=0.16\linewidth]{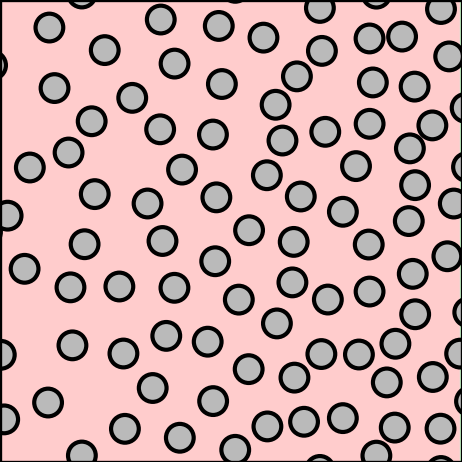} & 
    \includegraphics[width=0.16\linewidth]{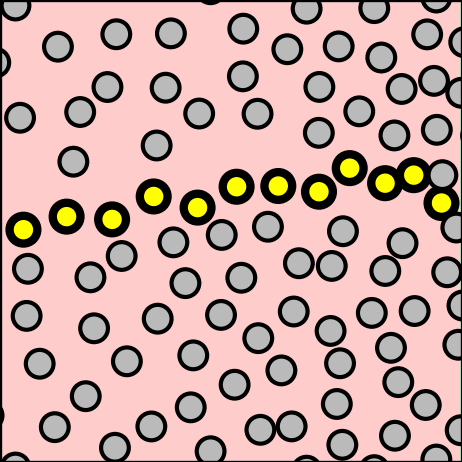} &
    \includegraphics[width=0.16\linewidth]{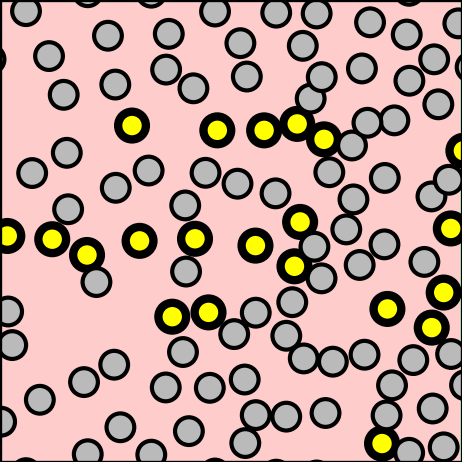} &
    \includegraphics[width=0.16\linewidth]{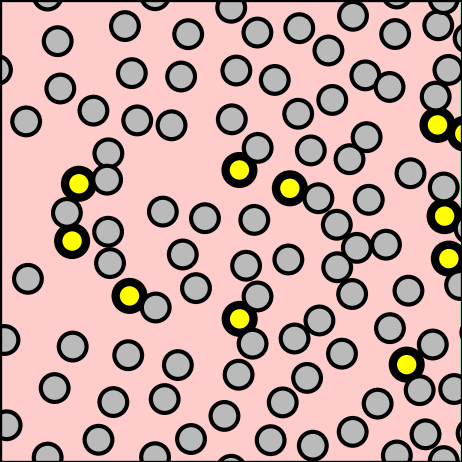} &
    \includegraphics[width=0.16\linewidth]{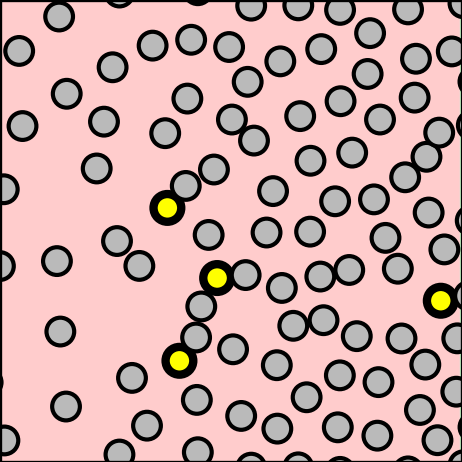} &
    \includegraphics[width=0.16\linewidth]{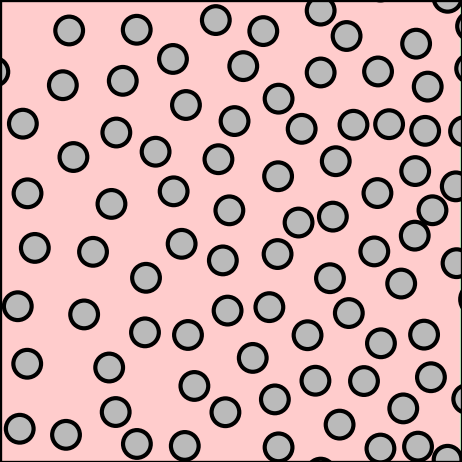} \\ 
    \small (a) 0.0\,s & \small (b) 18.0\,s & \small (c) 265.0\,s & \small (d) 306.0\,s & \small (e) 321.2\,s & \small (f) 367.5\,s
    \end{tabular}
    \endgroup
    \caption{Snapshots of region $\mathcal{W}_C \setminus \mathcal{W}_G$ from a typical trial with 200 robots navigating through a counter-flow crowd of density 0.30. The final snapshot was taken once all robots reached the goal. Top: Adaptive platoon strategy; bottom: greedy strategy.}
    \label{fig:snapshots}
\end{figure}

Fig.~\ref{fig:snapshots} shows a sequence of snapshots where $200$ robots use the adaptive platoon and greedy strategies, respectively, both formations being initialized as a line.
It reveals a path akin to a ``highway'' created by the adaptive platoon strategy, facilitating the passage of subsequent robots through the crowd region while minimizing disruption. The greedy strategy instead causes the robots to spread out. By not taking advantage of each others' efforts, the robots achieve a reduced navigation speed while causing increased disruption to the crowd.

\section{Conclusion}
\label{Conc}

This paper explored the potential benefits of platooning, and less constrained, greedy strategies, for navigating crowded environments. 
The effectiveness of the strategies was examined using crowds that adhere to the social force model, and are passive, moving against the flow of the robots, or perpendicularly to it. 

All controllers succeeded in traversing through the crowd for a wide range of crowd densities. In low-density settings, the greedy strategies reached the goal faster than platoon strategies. In high-density settings, the platooning strategies reached the goal consistently faster in all but perpendicular-flow crowd scenarios.
We quantified the level of disruption that the robots caused to the crowd, revealing that the platoon strategy is consistently the least disruptive in the passive and counter-flow crowd scenarios, whereas it is the most disruptive in the perpendicular-flow crowd scenario.
We proposed an adaptive platoon strategy that enabled platoon formations to split and merge dynamically, and showed that it successfully combines the advantages of the platoon and greedy strategies on either performance metric.

Future work will investigate more elaborate platoon controllers (e.g. based on \cite{morbidi2011visibility}) to improve performance and safety by considering input constraints, and will validate the findings using real robots and human participants.




\section*{Acknowledgment.} JAG's research is sponsored by a PhD scholarship from the government of Panama via IFARHU-SENACYT.

This project has received funding from the European Union's Horizon Europe Framework Programme under Grant Agreement No. 101093046.

\bibliographystyle{styles/bibtex/splncs03}
\bibliography{author.bib}

%







\end{document}